\documentclass[journal,twoside,print]{ieeecolor}
\usepackage{generic}
\providecommand{\refname}{REFERENCES}
\usepackage{etoolbox}
\makeatletter
\patchcmd{\thebibliography}
  {\itemsep 0pt plus pt\relax}
  {\itemsep 0pt plus 0.5pt\relax}
  {}
  {\PackageWarning{KGR}{Could not patch bibliography item spacing}}
\makeatother
\usepackage{cite}
\usepackage{amsmath,amssymb,amsfonts}
\usepackage{algorithmic}
\usepackage{graphicx}
\usepackage{algorithm,algorithmic}
\usepackage{hyperref}
\hypersetup{hidelinks=true}
\usepackage{textcomp}
\usepackage{multirow}
\usepackage{stfloats}
\makeatletter
\@ifundefined{IEEEproof}{%
  \newenvironment{IEEEproof}{\par\noindent\textit{Proof:}\ \ignorespaces}{\hfill$\blacksquare$\par\vspace{2pt}}%
}{}
\makeatother
\def\BibTeX{{\rm B\kern-.05em{\sc i\kern-.025em b}\kern-.08em
    T\kern-.1667em\lower.7ex\hbox{E}\kern-.125emX}}
\makeatletter
\def\ps@arxivheadings{%
  \def\@oddhead{\hbox{}\hfil\thepage}%
  \def\@evenhead{\thepage\hfil\hbox{}}%
  \def\@oddfoot{}%
  \def\@evenfoot{}%
}
\makeatother
\begin{document}
\title{K-space Gaussian Representation for Parallel MRI Reconstruction}
\author{Yu Guan, Mingyu Hu, Jiale Hu, Zhuoxu Cui, Dong Liang, \IEEEmembership{Senior Member, IEEE},\\ and Qiegen Liu, \IEEEmembership{Senior Member, IEEE}
\thanks{This work was supported in part by the National Natural Science Foundation of China under Grant 62122033 and Grant 62161027, and in part by the Natural Science Foundation of Jiangxi Province under Grant 20262BAC200309. (Y. Guan and M. Hu contributed equally to this work. Corresponding author: Qiegen Liu.)}
\thanks{Y. Guan is with the School of Advanced Manufacturing and the School of Information Engineering, Nanchang University, Nanchang 330031, China, and also with the Key Laboratory of Biomedical Imaging Science and System, Chinese Academy of Sciences, Shenzhen 518055, China (e-mail: guanyu@email.ncu.edu.cn).}
\thanks{M. Hu, J. Hu, and Q. Liu are with the School of Information Engineering, Nanchang University, Nanchang 330031, China (e-mail: humingyu0131@gmail.com, 17320067309@163.com, liuqiegen@ncu.edu.cn).}
\thanks{Z. Cui and D. Liang are with the Research Center for Medical AI, Shenzhen Institutes of Advanced Technology, Chinese Academy of Sciences, Shenzhen 518055, China, and also with the Key Laboratory of Biomedical Imaging Science and System, Chinese Academy of Sciences, Shenzhen 518055, China (e-mail: zhuoxu.cui@siat.ac.cn, dong.liang@siat.ac.cn).}}

\maketitle
\pagestyle{arxivheadings}
\thispagestyle{arxivheadings}

\begin{abstract}
Accelerated magnetic resonance imaging (MRI) aims to recover the k-space signal from acquired measurements, where accurate estimation of missing samples is essential for high-fidelity reconstruction. Existing k-space reconstruction methods estimate missing samples through interpolation operators or structure priors defined on discrete sampling grids. Although these formulations effectively exploit local interpolation relationships and global k-space redundancy, they reconstruct only discrete frequency coefficients and therefore do not explicitly model the underlying continuous signal. To overcome this limitation, we propose K-space Gaussian Representation (KGR), the first explicit continuous representation formulated directly in the native k-space domain. Rather than estimating unknown samples on discrete grids, KGR parameterizes the continuous signal using Gabor-Gaussian primitives with shared spatial geometry, yielding a compact representation that naturally preserves inter-coil correlations. Because unconstrained continuous fitting does not necessarily satisfy the intrinsic structural properties of multi-coil signal, the estimated representation is projected onto a low-rank manifold to enforce the algebraic constraints arising from smoothly varying phase and coil redundancy. A frequency-adaptive fitting strategy accommodates the heterogeneous characteristics of different k-space regions. Comprehensive validation across multiple datasets and sampling schemes shows consistent improvements over representative reconstruction baselines in both quantitative metrics and visual quality. These results suggest that explicit continuous parameterization of native k-space provides a principled framework for integrating continuous signal modeling with structured low-rank reconstruction.
\end{abstract}


% Keep the IEEE Introduction heading with its drop-cap opening paragraph.
\newpage
\section{Introduction}
\IEEEPARstart{M}{agnetic} resonance imaging (MRI) provides excellent soft-tissue contrast without ionizing radiation, yet prolonged acquisition time remains a major limitation in many clinical applications \cite{pruessmann1999sense}. Parallel MRI addresses this challenge by exploiting the complementary spatial encoding provided by multiple receiver coils, enabling accelerated acquisition with reduced sampling density \cite{griswold2002grappa}. However, acceleration inevitably leaves missing k-space samples, requiring accurate recovery of frequency coefficients while preserving the underlying signal structure. Therefore, the performance of parallel MRI reconstruction critically depends on how the unacquired k-space signal is represented and constrained during reconstruction.

\begin{figure}[H]
    \centering
    \includegraphics[width=\columnwidth]{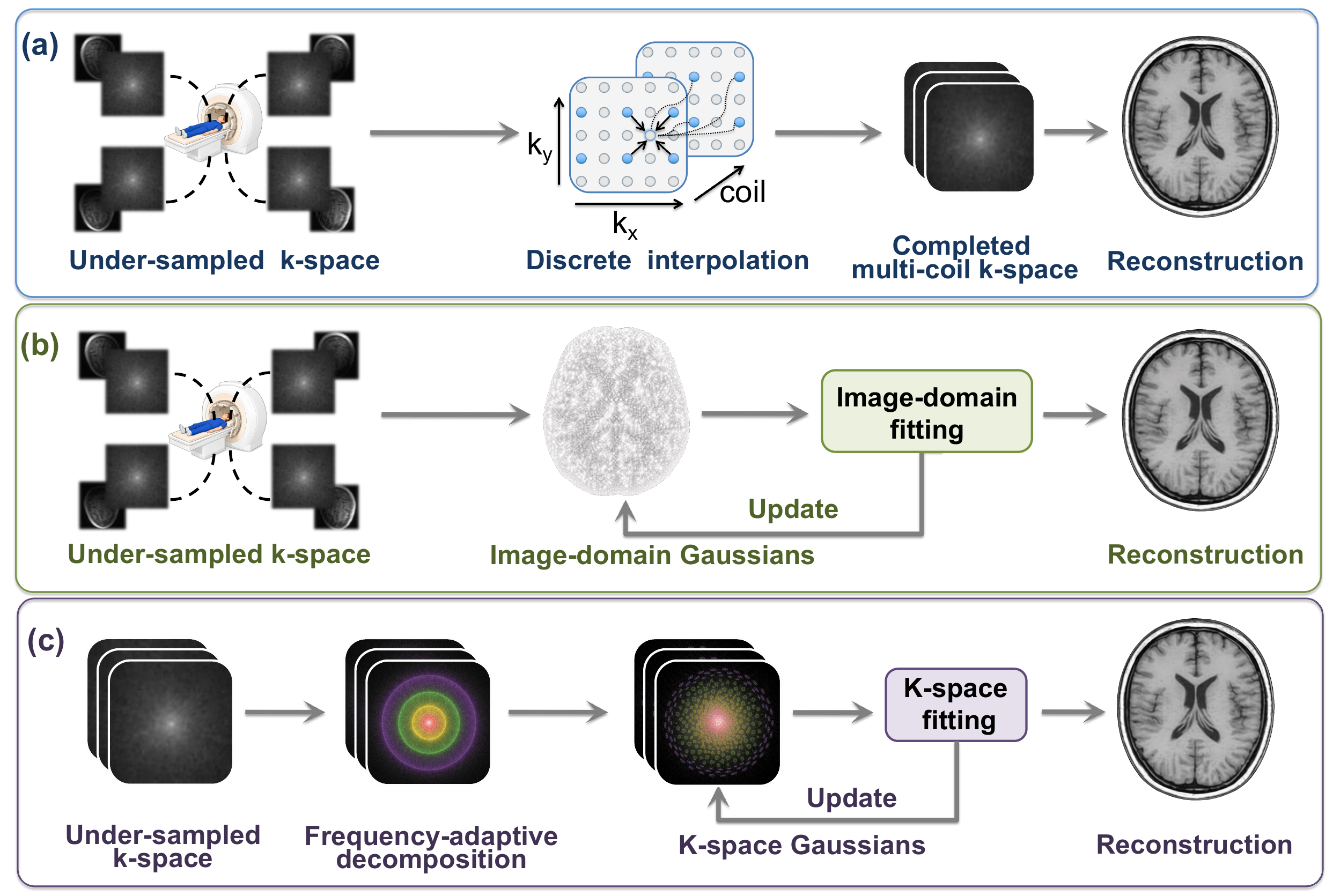}
    \caption{Conceptual comparison of representations for parallel MRI reconstruction. (a) Discrete k-space reconstruction estimates missing samples on a discrete grid. (b) Image-domain Gaussian representation places primitives in the spatial domain. (c) K-space Gaussian Representation (KGR) places the Gaussian representation in native k-space and models missing samples through frequency-adaptive decomposition.}
    \label{fig:motivation}
\end{figure}

Existing methods employ different strategies to model missing multi-coil k-space signal. Conventional interpolation-based approaches recover unacquired samples on the discrete Fourier grid using local kernels or calibration relationships, as exemplified by GRAPPA \cite{griswold2002grappa}. Structured low-rank methods exploit the redundancy of multi-coil acquisitions by lifting local k-space neighborhoods into block-Hankel matrices and enforcing low-rank constraints \cite{shin2014sake}. More recently, learning-based approaches have introduced data-driven priors through neural networks to learn reconstruction mappings or iterative update rules \cite{akcakaya2019raki,yaman2020ssdu,liang2020inverse}. Despite their effectiveness, these approaches fundamentally represent k-space either as discrete samples, structured matrices or implicit network parameters, rather than as an explicit continuous signal.

This observation reveals an unexplored representation gap in accelerated parallel MRI. The measured k-space signal originates from a continuous Fourier encoding process, whereas existing methods operate on discrete sampling locations or lifted algebraic structures \cite{shin2014sake,jin2016hankel}. Conversely, continuous representations have been primarily developed for image-domain or volumetric signals, where the representation is constructed after Fourier inversion or encoded implicitly through coordinate-based neural functions \cite{sitzmann2020siren}. As illustrated in Fig.~\ref{fig:motivation}, an explicit continuous representation defined directly in multi-coil k-space could provide a complementary perspective by modeling the underlying frequency signal before missing samples are recovered.

Structured low-rank modeling remains an effective prior for parallel MRI reconstruction because multi-coil k-space exhibits local redundancy induced by smoothly varying phase and coil sensitivity correlations \cite{haldar2014loraks,haldar2016ploraks}. However, the Hankel low-rank formulation characterizes discrete relationships among sampled neighborhoods and does not explicitly describe continuous variations across the frequency domain. A natural question arises: Can a continuous k-space representation be developed while retaining compatibility with the structured low-rank property of multi-coil acquisitions?

Gaussian representations provide an explicit parameterization strategy for continuous signal modeling \cite{park1991universal,kerbl20233dgs}. Recent advances inspired by 3D Gaussian splatting have demonstrated their potential in medical inverse problems, including MRI reconstruction \cite{zha2024r2gaussian,terpstra2026fast}. Nevertheless, existing Gaussian formulations mainly focus on image-domain or volumetric reconstruction, leaving their application to native complex k-space signal unexplored. Extending Gaussian representation from image space to the k-space domain may enable a new reconstruction paradigm that combines continuous frequency modeling with physics-driven k-space constraints.

To address this gap, we propose \textbf{K}-space \textbf{G}aussian \textbf{R}epresentation (KGR), an explicit continuous parameterization of k-space signal for accelerated parallel MRI reconstruction. Unlike conventional approaches that estimate missing Fourier coefficients only on discrete grids, KGR models the underlying k-space signal using frequency-localized Gabor–Gaussian primitives with shared geometric parameters across coils. This formulation enables continuous characterization of local spectral variations while preserving inter-coil relationships. Furthermore, KGR is integrated with Hankel low-rank refinement to enforce global multi-coil consistency and maintain compatibility with established k-space priors. The proposed optimization framework does not require pretrained reconstruction networks, providing a direct measurement-driven solution for accelerated parallel MRI. 

The main contributions are summarized as follows:
\begin{itemize}
    \item \textbf{\textit{We introduce a continuous signal representation framework for complex k-space signal reconstruction.}} Unlike conventional approaches that recover missing pixel on discrete grids or rely on implicit representations, the proposed KGR explicitly models the underlying k-space signal using localized Gaussian basis functions.
    \item \textbf{\textit{We develop a physics-compatible scheme that integrates continuous k-space modeling with low-rank constraints.}} By combining frequency-adaptive Gaussian fitting with Hankel structured low-rank refinement, KGR simultaneously captures local spectral variations and global multi-coil correlations in the acquired k-space.
    \item \textbf{\textit{We provide an analysis of the compatibility between k-space representation and low-rank constraints.}} The derived error analysis characterizes how representation accuracy and low-rank consistency affect reconstruction fidelity, offering insights into the design of continuous signal models for accelerated MRI.
\end{itemize}

\section{Related Work}

\subsection{K-space Reconstruction}

Parallel MRI measures multiple coil-weighted views of the same object. For coil $c$, the acquired k-space samples are commonly modeled as:
\begin{equation}
    \hat{K}_c = M_{\Omega} \odot \mathcal{F}\{s_c \rho\} + \epsilon_c,
    \quad c=1,\cdots,C,
    \label{eq:parallel_mri_signal}
\end{equation}
where $C$ is the total number of receiver coils, $\rho$ denotes the object, $s_c$ is the coil-dependent modulation, $\mathcal{F}$ is the Fourier transform, $M_{\Omega}$ is the sampling mask supported on the acquired index set $\Omega$, and $\epsilon_c$ denotes measurement noise.

Under accelerated sampling, the inverse problem is to recover the missing multi-coil k-space from the acquired samples. A generic regularized k-space reconstruction problem can be written as:
\begin{equation}
    K_{\mathrm{rec}} = \arg\min_{K}
    \sum_{c=1}^C
    \left\|M_{\Omega}\odot K_c - \hat{K}_c\right\|_2^2
    + \lambda \mathcal{R}(K),
    \label{eq:kspace_inverse_problem}
\end{equation}
where $K=\{K_c\}_{c=1}^C$ denotes the multi-coil k-space to be reconstructed and $\mathcal{R}(K)$ encodes a prior or consistency constraint. Different reconstruction methods can therefore be understood through their parameterization of $K$ and their choice of $\mathcal{R}$.

Native k-space reconstruction treats Fourier samples as unknown, allowing data consistency to be enforced directly at acquired locations. Classical examples include GRAPPA, which learns local interpolation kernels \cite{griswold2002grappa}, and SPIRiT, which enforces self-consistency over multi-coil k-space \cite{lustig2010spirit}. These methods highlight the appeal of the native frequency domain: Coil correlations and neighboring-frequency structure can be used before image formation. However, their representations remain tied to discrete samples, kernels, or consistency equations, so local frequency-domain variation is exploited through grid-based relations rather than exposed as an explicit parameterization before sample recovery. From a k-space reconstruction perspective, a useful continuous representation should preserve native data consistency and still interact with structured multi-coil consistency \cite{huang2026gaborprimitives}. This is the setting in which a k-space Gaussian representation becomes meaningful: It asks whether frequency-localized primitives can supply the missing continuous parameterization while staying in the same domain as classical k-space methods.

\subsection{Low-Rank Modeling of K-space}

Beyond interpolation and self-consistency, undersampled MRI is commonly regularized by priors on images, k-space, or learned reconstruction maps. Compressed sensing exploits transform-domain sparsity \cite{lustig2007sparse}, whereas structured low-rank methods enforce constraints derived from finite spatial support, smoothly varying phase, and inter-coil correlation through annihilation relations in overlapping k-space neighborhoods \cite{shin2014sake}. Let $\mathcal{E}_n$ extract the $n$-th local multi-coil k-space neighborhood, with $d$ denoting its vectorized dimension. A generic block-Hankel operator can be written as:
\begin{equation}
\begin{aligned}
    \mathcal{H}(K) =
    \begin{bmatrix}
    (\mathcal{E}_1K)^\top \\
    \vdots \\
    (\mathcal{E}_NK)^\top
    \end{bmatrix}
    \in \mathbb{C}^{N\times d}, \\
    \operatorname{rank}(\mathcal{H}(K)) \ll \min(N,d).
\end{aligned}
\label{eq:structured_hankel_prior}
\end{equation}
SAKE \cite{shin2014sake}, LORAKS \cite{haldar2014loraks}, ALOHA \cite{lee2016aloha}, and annihilating-filter formulations \cite{jin2016hankel} instantiate this idea through different lifting, completion, and constraint designs. These structured approaches support calibrationless multi-coil reconstruction by enforcing low-rank structure in lifted neighborhood matrices, yet they do not parameterize the continuous frequency variation underlying native k-space samples.

\subsection{Learning-based K-space Methods}

Learning-based reconstruction methods improve adaptivity through learned priors, proximal mappings, or interpolation rules. Variational networks and model-based deep unrolling incorporate learned regularization \cite{hammernik2018variational,aggarwal2019modl}, while scan-specific approaches such as RAKI learn k-space interpolation directly from the acquired scan \cite{akcakaya2019raki}. Self-supervised methods reduce dependence on fully sampled training targets \cite{yaman2020ssdu,knoll2020survey} and extend to coordinate-based continuous functions fitted to non-Cartesian k-space \cite{huang2024neural}. Recent work such as PISCO demonstrates that neural implicit k-space models can benefit from self-consistency regularization inspired by parallel imaging in dynamic MRI \cite{spieker2025pisco}. These methods are valuable because they adapt the reconstruction model to acquisition-specific data. However, their continuous representation is encoded in learned operators or network weights. Coordinate networks define continuous functions, but frequency support, orientation, and coil coupling are encoded through weights rather than exposed as primitive parameters \cite{sitzmann2020siren,shen2026imjplus}. This motivates a complementary explicit representation. KGR uses Gabor-Gaussian primitives whose centers, covariances, and coil-specific amplitudes make frequency-domain structure accessible.

\subsection{Gaussian Representations in Medical Imaging}

Gaussian representations model continuous signals with explicit primitives, making amplitudes, centers, and covariances the primary variables instead of pixels or network weights. In its basic form, a Gaussian field $f(\mathbf{x})$ can be written as:
\begin{equation}
    f(\mathbf{x}) = \sum_{j=1}^J \alpha_j
    \exp\!\left[-\frac{1}{2}(\mathbf{x}-\boldsymbol{\mu}_j)^\top
    \Sigma_j^{-1}(\mathbf{x}-\boldsymbol{\mu}_j)\right],
    \label{eq:gaussian_field_prior}
\end{equation}
where each primitive has an amplitude $\alpha_j$, center $\boldsymbol{\mu}_j$, and covariance $\Sigma_j$. This explicit-primitive view was popularized in modern radiance-field rendering by 3D Gaussian splatting \cite{kerbl20233dgs} and has since been adapted to inverse medical imaging. Applications have included CT and ultrasound reconstruction \cite{gao2024ddgsct,eid2025ultragauss,noh2025signedgaussians,liang2025innergs}, as well as MRI tasks such as dynamic reconstruction \cite{terpstra2026fast}, super-resolution \cite{liu2026physicsdriven,jo2026pingsx}, and multi-stack reconstruction \cite{zheng2026mgaussian}. These works generally parameterize image-domain intensities, volumes, or spatial latent variables and evaluate them through the corresponding imaging forward model \cite{peng2026threedimensional}.

Parallel MRI has a different representation target: The measured variables are complex multi-coil k-space samples. In this domain, acquired samples define data-consistency constraints, while frequency-dependent energy decay, local directional correlation, and coil coupling are part of the signal structure. Recent Gabor primitive formulations show that Gaussian envelopes can be modulated to improve high-frequency MRI representation \cite{huang2026gaborprimitives}. However, the Gaussian medical imaging literature cited above has mainly treated Gaussian primitives as representations of image-space objects or latent spatial variables, including volumetric and dynamic MRI settings \cite{liu2026physicsdriven,zheng2026mgaussian,jo2026pingsx}. Native complex multi-coil k-space has remained less explored as a domain for explicit continuous Gaussian representation. The question is whether primitives can be formulated directly in k-space while retaining data consistency at acquired samples and the structured multi-coil consistency used by calibrationless k-space methods \cite{lee2016aloha}.

\section{Methodology}

\subsection{Motivation}

\begin{figure}[!b]
    \centering
    \includegraphics[width=0.8\columnwidth]{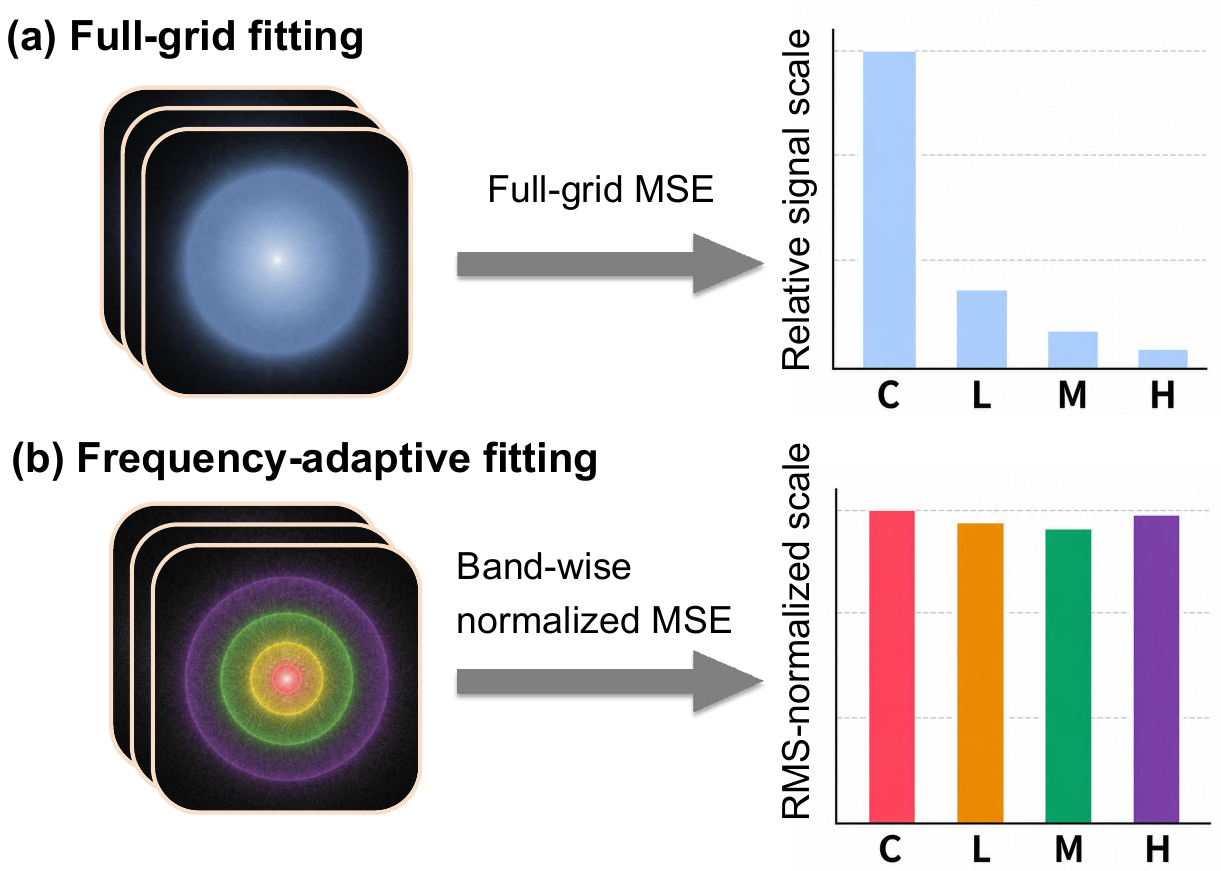}
    \caption{Frequency-adaptive Gaussian fitting. (a) A full-grid loss can be biased toward the high-energy central k-space. (b) KGR computes a mean-squared residual separately within the central (C), low (L), mid (M), and high (H) bands and normalizes by squared band RMS.}
    \label{fig:frequency_adaptive_motivation}
\end{figure}

KGR directly represents the complex k-space signal with continuous Gaussian primitives. The first difficulty is its large frequency-dependent dynamic range: The central region has a much larger magnitude than peripheral regions. A single full-grid loss can therefore bias optimization toward the high-energy center and underweight weaker high-frequency components. We instead compute a mean-squared residual within each frequency band and normalize it by the squared band RMS magnitude $\eta_r^2$. As illustrated in Fig.~\ref{fig:frequency_adaptive_motivation}, this converts absolute residuals into band-wise relative errors and reduces the effect of frequency-dependent signal magnitude.

\begin{figure}[!t]
    \centering
    \includegraphics[width=0.92\columnwidth]{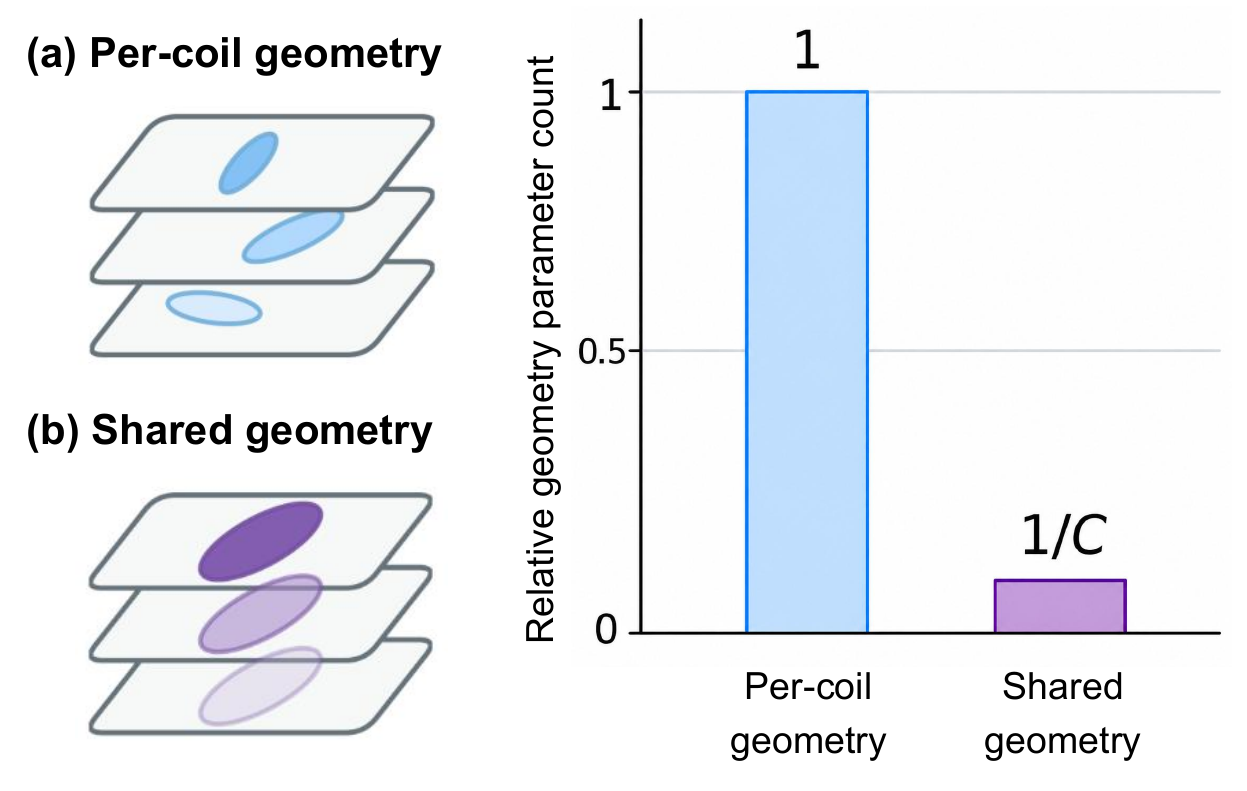}
    \caption{Cross-coil sharing of Gaussian geometry. (a) Per-coil geometry uses an independent set of primitive centers, covariances, and carriers for each receiver. (b) KGR shares one geometry set across $C$ coils while retaining coil-specific complex amplitudes. For the same number of primitives available to each coil, the normalized geometry parameter count decreases from $1$ to $1/C$.}
    \label{fig:cross_coil_geometry}
\end{figure}

Frequency balancing addresses the first problem, but parallel MRI creates two more. First, assigning independent Gaussian geometry to each of $C$ receiver coils duplicates the primitive centers, covariances, and carriers $C$ times, which can make a band-specific representation unnecessarily large. Because all coils encode the same object under receiver-dependent modulation, KGR shares one geometry set across coils while retaining coil-specific complex amplitudes. For the same number of primitives per coil, the normalized geometry parameter count decreases from $1$ to $1/C$, as shown in Fig.~\ref{fig:cross_coil_geometry}. This reduction applies only to the geometry parameters, the complex amplitudes remain coil-specific.

\begin{figure*}[!t]
    \centering
    \includegraphics[width=\textwidth]{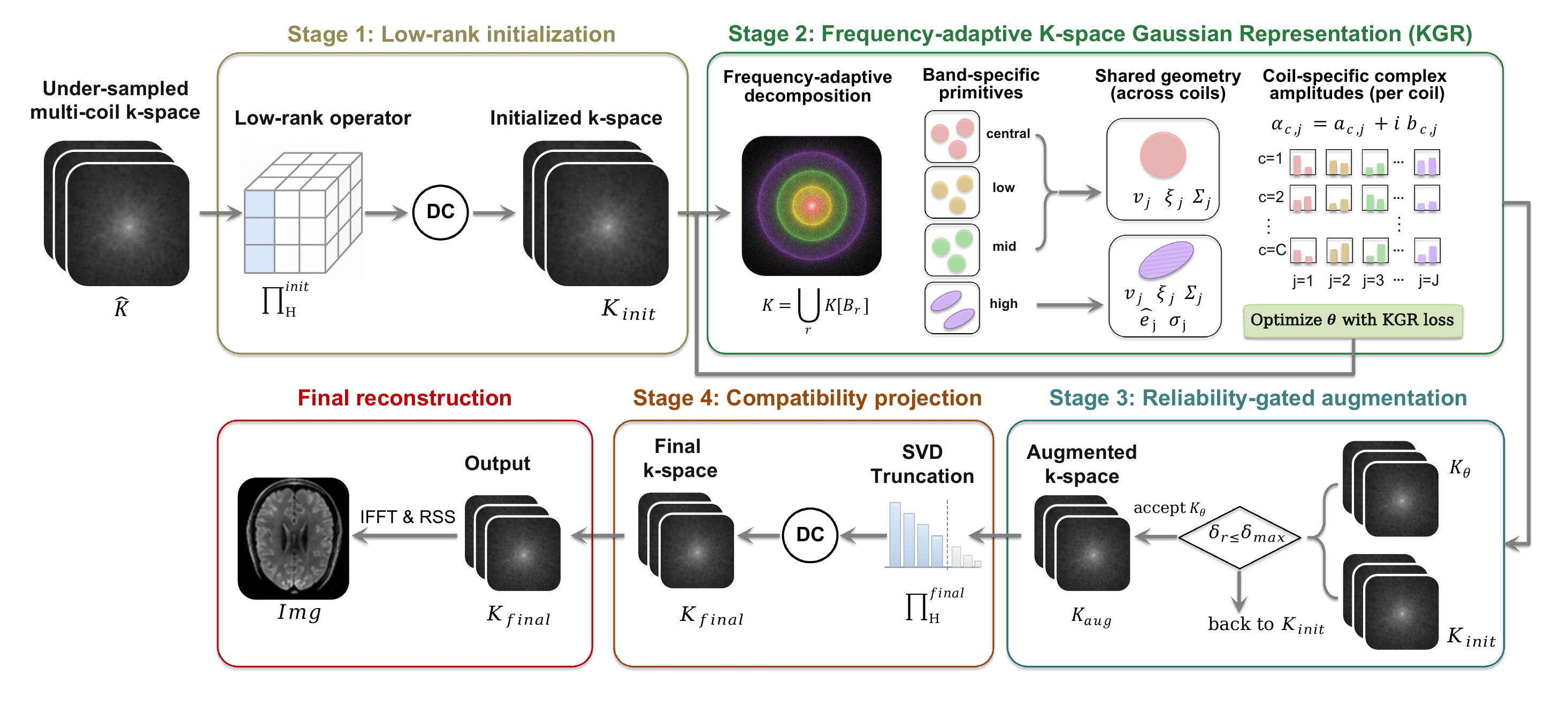}
    \caption{Overview of the proposed KGR reconstruction pipeline. The undersampled multi-coil k-space first undergoes a Hankel projection with acquired-sample replacement to obtain a low-rank initialization. Frequency-adaptive KGR then fits band-specific Gaussian and Gabor-Gaussian primitives with shared geometry and coil-specific complex amplitudes to predict $K_{\theta}$. A reliability-gated augmentation accepts reliable KGR predictions and retains the initialization elsewhere. A final Hankel low-rank compatibility projection produces $K_{\mathrm{final}}$ before root-sum-of-squares (RSS) visualization. The data-consistency (DC) blocks replace values at acquired locations $\Omega$.}
    \label{fig:kgr_pipeline}
\end{figure*}

These designs adapt Gaussian primitives to the frequency imbalance, parameter growth, and local oscillation of complex multi-coil k-space. Frequency-adaptive Gaussian fitting captures local spectral variations, but its local primitives do not explicitly preserve correlations across receiver coils and neighboring k-space locations. We therefore pair KGR with Hankel structured low-rank refinement, which enforces these global multi-coil correlations. A low-rank reconstruction initializes Gaussian fitting, a reliability gate accepts KGR predictions only where they agree with acquired data, and a final low-rank projection restores global consistency. Fig.~\ref{fig:kgr_pipeline} summarizes the complete procedure.

\subsection{K-space Gaussian Representation}

We denote the fully sampled multi-coil k-space signal of a single scan by $K^* = \{K_c^*\}_{c=1}^C \in \mathbb{C}^{C \times H \times W}$, where $C$ receiver coils are sampled on an $H\times W$ Cartesian k-space grid. The acquired measurements are denoted by $\hat{K}$, as described in Eq.~\eqref{eq:parallel_mri_signal}. We estimate $K^*$ from $\hat{K}$ by constructing a native continuous representation of the complex multi-coil k-space signal and enforcing compatibility with the low-rank prior.

KGR models multi-coil k-space as a continuous function of the k-space coordinate $\mathbf{k}=(k_x,k_y)^\top$. We define the $j$-th k-space Gabor-Gaussian primitive as:
\begin{equation}
    \phi_j(\mathbf{k}) =
    \exp\!\left[-\frac{1}{2}\mathbf{d}_j^\top\Sigma_j^{-1}\mathbf{d}_j\right]
    \exp\!\left(i\boldsymbol{\xi}_j^\top\mathbf{d}_j\right),
    \quad \mathbf{d}_j=\mathbf{k}-\boldsymbol{\nu}_j,
    \label{eq:gabor_atom}
\end{equation}
with primitive center $\boldsymbol{\nu}_j \in \mathbb{R}^2$, positive-definite covariance $\Sigma_j \in \mathbb{R}^{2\times2}$ controlling the envelope shape, and learnable sinusoidal carrier $\boldsymbol{\xi}_j \in \mathbb{R}^2$ controlling local complex modulation. This Gabor-Gaussian form follows recent high-frequency Gaussian representations \cite{watanabe2026neuralgabor,huang2026gaborprimitives}.

\begin{figure}[!t]
    \centering
    \includegraphics[width=0.8\columnwidth]{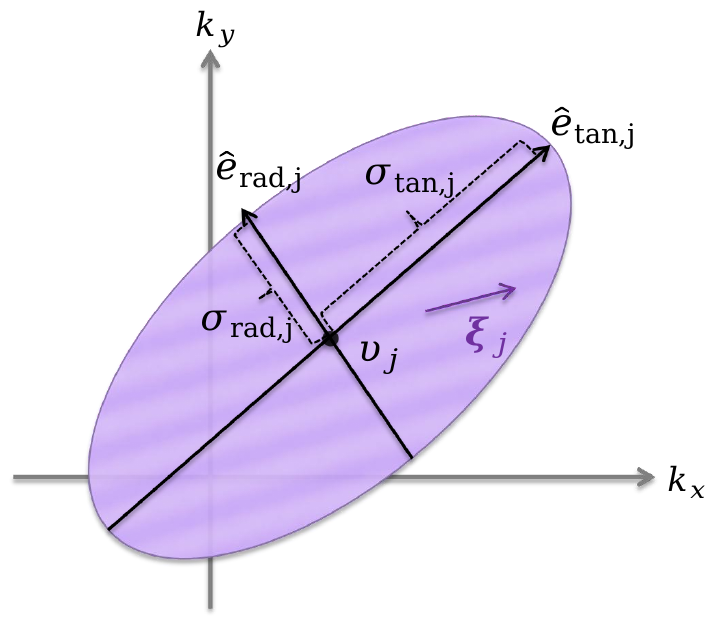}
    \caption{The primitive is centered at $\boldsymbol{\nu}_j$ and uses a Gabor carrier $\boldsymbol{\xi}_j$ inside an anisotropic Gaussian envelope. In the high-frequency band, the covariance frame is aligned with the local radial and tangential directions $\hat{\mathbf{e}}_{\mathrm{rad},j}$ and $\hat{\mathbf{e}}_{\mathrm{tan},j}$, with widths $\sigma_{\mathrm{rad},j}$ and $\sigma_{\mathrm{tan},j}$.}
    \label{fig:gabor_gaussian_atom}
\end{figure}

The shared geometric parameters of a k-space primitive are visualized in Fig.~\ref{fig:gabor_gaussian_atom}. The covariance can be specialized to isotropic or directionally anisotropic forms. The central, low-, and mid-frequency regions use an isotropic covariance $\Sigma_j=\sigma_j^2 I$, while high-frequency k-space uses an anisotropic covariance in the local radial-tangential frame. We define $\hat{\mathbf{e}}_{\mathrm{rad},j}=\boldsymbol{\nu}_j/\|\boldsymbol{\nu}_j\|_2$ as the radial direction and $\hat{\mathbf{e}}_{\mathrm{tan},j}$ as its orthogonal tangential direction. The covariance eigenvectors are set to $\{\hat{\mathbf{e}}_{\mathrm{rad},j},\hat{\mathbf{e}}_{\mathrm{tan},j}\}$ with eigenvalues $\sigma_{\mathrm{rad},j}^2$ and $\sigma_{\mathrm{tan},j}^2$. When $\sigma_{\mathrm{tan},j}>\sigma_{\mathrm{rad},j}$, high-frequency primitives extend along constant-radius directions and remain radially localized. Coil-specific modulation is introduced separately through the complex amplitudes in the multi-coil expansion below.

The predicted value of coil $c$ is:
\begin{equation}
    K_{\theta,c}(\mathbf{k}) =
    \sum_{r=1}^4\sum_{j\in\mathcal{J}_r}
    (a_{c,j} + i b_{c,j})\phi_j(\mathbf{k}),
    \label{eq:kgr_multicoil}
\end{equation}
where $\mathcal{J}_r$ indexes the primitives assigned to frequency band $r$, and $a_{c,j}$ and $b_{c,j}$ are coil-specific real and imaginary amplitudes. KGR shares $\{\boldsymbol{\nu}_j,\Sigma_j,\boldsymbol{\xi}_j\}$ across coils and learns complex amplitudes per receiver. This factorization couples the receivers through common frequency-domain support without explicit sensitivity maps, while the amplitudes preserve receiver-dependent modulation.

The resulting continuous representation is fitted to a low-rank initialization and subsequently returned to the structured low-rank set, as detailed in the following subsections.
\vspace{-0.8\baselineskip}
\subsection{Frequency-Adaptive KGR Fitting}

Before fitting KGR, we obtain a full-grid estimate through the low-rank projection:
\begin{equation}
    K_{\mathrm{init}} = \Pi_{\mathrm{H}}(\hat{K}),
    \label{eq:init_projection}
\end{equation}
where $\Pi_{\mathrm{H}}$ denotes a Hankel low-rank projection with acquired-sample replacement. KGR fits $K_{\theta}$ to $K_{\mathrm{init}}$ while remaining anchored to measured samples; the reliability gate defined below determines where its predictions enter $K_{\mathrm{aug}}$.

We divide the grid into four radial bands according to $\|\mathbf{k}\|_2$: Central, low, mid, and high frequency. For band $r$, $\mathcal{B}_r$ denotes its set of grid points and $\Omega_r=\Omega\cap\mathcal{B}_r$ denotes its acquired subset. We use $\mathbb{E}_{\mathcal{A}}$ to denote the arithmetic mean over all coils and k-space locations in a set $\mathcal{A}$. KGR parameters are optimized by combining a representation-fitting term with an acquired-sample anchoring term:
\begin{equation}
\begin{aligned}
    \mathcal{L}_{\mathrm{KGR}}(\theta)
    &= \sum_{r=1}^4
    \left(\mathcal{L}_{\mathrm{fit}}^{(r)}
    + \lambda_{\Omega}\mathcal{L}_{\mathrm{data}}^{(r)}\right), \\
    \mathcal{L}_{\mathrm{fit}}^{(r)}
    &= \frac{\mathbb{E}_{\mathcal{B}_r}
    \left[\left|K_{\theta}-K_{\mathrm{init}}\right|^2\right]}{\eta_r^2}, \\
    \mathcal{L}_{\mathrm{data}}^{(r)}
    &= \frac{\mathbb{E}_{\Omega_r}
    \left[\left|K_{\theta}-\hat{K}\right|^2\right]}{\eta_r^2}.
\end{aligned}
    \label{eq:kgr_loss}
\end{equation}
Here $\eta_r$ is the RMS magnitude of $K_{\mathrm{init}}$ over coils and acquired locations in $\Omega_r$. The fitting term $\mathcal{L}_{\mathrm{fit}}^{(r)}$ distills the low-rank initialization into the continuous parameterization, including unsampled locations where no direct measurement is available. Since the initialization is itself an estimate, $\mathcal{L}_{\mathrm{data}}^{(r)}$ separately anchors the representation to the measurements, with $\lambda_{\Omega}$ setting the anchor strength. Taking the mean removes dependence on the number of entries in each band, while division by $\eta_r^2$ converts the residual into a band-wise relative MSE. This normalization reduces deterministic scale differences without forcing the realized losses of all bands to be equal.

After fitting, KGR predictions are used only in frequency bands where the fitted representation remains consistent with the acquired data. For each band $r$, we compute the relative MSE at acquired positions:
\begin{equation}
    \delta_r =
    \frac{\mathbb{E}_{\Omega_r}
    \left[\left|K_{\theta}-\hat{K}\right|^2\right]}{\eta_r^2}.
    \label{eq:band_residual}
\end{equation}
This band-wise relative MSE gates whether KGR predictions are trusted in that band. The augmented k-space is then constructed by accepting KGR fits only where $\delta_r$ remains below threshold:
\begin{equation}
K_{\mathrm{aug}}[c,\mathbf{k}] =
\begin{cases}
\hat{K}[c,\mathbf{k}], & \mathbf{k}\in\Omega, \\
K_{\theta,c}(\mathbf{k}), & \mathbf{k}\notin\Omega,\; \delta_{r(\mathbf{k})}\leq \delta_{\mathrm{max}}, \\
K_{\mathrm{init}}[c,\mathbf{k}], & \mathrm{otherwise},
\end{cases}
\label{eq:gated_fill}
\end{equation}
where $r(\mathbf{k})$ denotes the frequency band containing $\mathbf{k}$ and $\delta_{\mathrm{max}}$ is a reliability threshold. The gate preserves data consistency at acquired positions and prevents unreliable KGR fits from replacing the low-rank initialization in unsampled regions. Hard replacement, rather than weighted blending, keeps the Gaussian completion as a distinct representation before the final low-rank projection.

\subsection{Low-Rank Compatibility Projection}

The reliability-gated KGR estimate is evaluated with the structured prior used during initialization and LR-GR, implemented by $\Pi_{\mathrm{H}}$ as in Eq.~\eqref{eq:structured_hankel_prior}. For a window size $w_h\times w_w$, denote the multi-coil block-Hankel lifting by $\mathcal{H}_{w_h,w_w}$ and its normalized adjoint, which averages overlapped patch estimates back to the k-space grid, by $\mathcal{H}_{w_h,w_w}^\dagger$. Given an input k-space estimate, one projection iteration alternates singular value thresholding (SVT), overlap-add aggregation, and acquired-sample replacement:
\begin{equation}
\begin{aligned}
    \tilde{K}^{(t+1)} &= \mathcal{H}_{w_h,w_w}^\dagger
    \left(\mathrm{SVT}_{\tau}\!\left(\mathcal{H}_{w_h,w_w}(K^{(t)})\right)\right), \\
    K^{(t+1)} &= \mathcal{P}_{\Omega}(\tilde{K}^{(t+1)}).
\end{aligned}
\label{eq:hankel_projection}
\end{equation}
Here $\mathrm{SVT}_{\tau}$ keeps singular values according to threshold $\tau$, and $\mathcal{P}_{\Omega}$ replaces acquired positions by the measured values $\hat{K}|_{\Omega}$. We implement $\Pi_{\mathrm{H}}$ as a generic structured low-rank Hankel projection of the form in Eq.~\eqref{eq:structured_hankel_prior} \cite{haldar2014loraks}. In the proposed method, this compatibility projection keeps the reconstruction aligned with measured samples and favors the Hankel low-rank structure shared across coils.

Using the representation $K_{\mathrm{aug}}$ produced by KGR and the projection operator $\Pi_{\mathrm{H}}$ defined above, LR-GR is obtained as:
\begin{equation}
    K_{\mathrm{final}} = \Pi_{\mathrm{H}}(K_{\mathrm{aug}}).
    \label{eq:lrgr_definition}
\end{equation}
Because the final step of Eq.~\eqref{eq:hankel_projection} is acquired-sample replacement, the finite-iteration implementation satisfies $M_{\Omega}\odot K_{\mathrm{final}}=\hat{K}$ whenever the last projection step is applied.

The analysis below isolates the condition under which the Gaussian stage benefits the low-rank step. KGR fitting targets this condition by replacing the purely discrete low-rank initialization only where a continuous Gaussian parameterization, anchored to acquired samples and screened by the reliability gate, gives a reliable completion. Under this premise, consider the ideal data-consistent Hankel-compatible set:
\begin{equation}
    \mathcal{C}_{\mathrm{H}} =
    \left\{K:
    M_{\Omega}\odot K=\hat{K},\;
    \operatorname{rank}\!\left(\mathcal{H}_{w_h,w_w}(K)\right)\leq q
    \right\},
    \label{eq:hankel_compatible_set}
\end{equation}
where $q$ denotes the target rank level represented by the selected thresholding rule. Let $\Pi_{\mathcal{C}_{\mathrm{H}}}$ denote an ideal best-approximation projection onto this set:
\begin{equation}
    \Pi_{\mathcal{C}_{\mathrm{H}}}(Z) \in
    \arg\min_{K\in\mathcal{C}_{\mathrm{H}}}\left\|K-Z\right\|_F .
    \label{eq:ideal_hankel_projection}
\end{equation}
The practical projection in Eq.~\eqref{eq:hankel_projection} is a finite-iteration approximation to this ideal compatibility operation.

\noindent\textbf{Proposition 1 (Error inheritance under Hankel compatibility projection).}
Assume that the target k-space $K^*$ belongs to $\mathcal{C}_{\mathrm{H}}$, and that the estimate augmented by KGR satisfies:
\begin{equation}
    \left\|K_{\mathrm{aug}}-K^*\right\|_F \leq \varepsilon_{\mathrm{G}}.
\end{equation}
Let $\bar{K}=\Pi_{\mathcal{C}_{\mathrm{H}}}(K_{\mathrm{aug}})$ be any ideal compatibility projection of $K_{\mathrm{aug}}$. Then the following hold:
\begin{itemize}
\item[(i)] The projection error is data consistent and is therefore supported on the unacquired set:
\begin{equation}
    M_{\Omega}\odot\left(\bar{K}-K^*\right)=0 .
\end{equation}
\item[(ii)] The error admits the decomposition bound:
\begin{equation}
    \left\|\bar{K}-K^*\right\|_F
    \leq
    \operatorname{dist}_F\!\left(K_{\mathrm{aug}},\mathcal{C}_{\mathrm{H}}\right)
    + \varepsilon_{\mathrm{G}}
    \leq 2\varepsilon_{\mathrm{G}},
    \label{eq:prop_decomposition}
\end{equation}
where $\operatorname{dist}_F(Z,\mathcal{C}_{\mathrm{H}})=\min_{K\in\mathcal{C}_{\mathrm{H}}}\|Z-K\|_F$ measures the structural incompatibility of $K_{\mathrm{aug}}$ with the Hankel low-rank set. The term $\varepsilon_{\mathrm{G}}$ quantifies the Gaussian approximation error relative to the target k-space.
\end{itemize}
\begin{IEEEproof}
(i) Both $\bar{K}$ and $K^*$ lie in $\mathcal{C}_{\mathrm{H}}$, so each satisfies $M_{\Omega}\odot K=\hat{K}$. Subtracting the two identities gives $M_{\Omega}\odot(\bar{K}-K^*)=0$, so the error vanishes at every acquired position and is carried entirely by the recovered samples.

(ii) Since $\bar{K}$ is a best approximation to $K_{\mathrm{aug}}$ in $\mathcal{C}_{\mathrm{H}}$, we have $\|K_{\mathrm{aug}}-\bar{K}\|_F=\operatorname{dist}_F(K_{\mathrm{aug}},\mathcal{C}_{\mathrm{H}})$. The triangle inequality then gives:
\begin{equation}
\begin{aligned}
    \left\|\bar{K}-K^*\right\|_F
    &\leq
    \left\|\bar{K}-K_{\mathrm{aug}}\right\|_F
    + \left\|K_{\mathrm{aug}}-K^*\right\|_F \\
    &= \operatorname{dist}_F\!\left(K_{\mathrm{aug}},\mathcal{C}_{\mathrm{H}}\right)
    + \left\|K_{\mathrm{aug}}-K^*\right\|_F .
\end{aligned}
\end{equation}
Because $K^*\in\mathcal{C}_{\mathrm{H}}$, the distance term satisfies $\operatorname{dist}_F(K_{\mathrm{aug}},\mathcal{C}_{\mathrm{H}})\leq\|K_{\mathrm{aug}}-K^*\|_F\leq\varepsilon_{\mathrm{G}}$. Substituting this bound into the previous inequality yields both the decomposition bound and the uniform bound $2\varepsilon_{\mathrm{G}}$.
\end{IEEEproof}

Proposition 1 gives a condition under which the Gaussian stage can benefit the low-rank step. Part (ii) decomposes the projected error into the approximation error $\varepsilon_{\mathrm{G}}$ of the continuous Gaussian representation and the structural incompatibility $\operatorname{dist}_F(K_{\mathrm{aug}},\mathcal{C}_{\mathrm{H}})$ with the Hankel low-rank set. KGR targets the approximation term by fitting acquired samples and neighboring frequency-domain structure, while the compatibility projection reduces structural mismatch by returning the estimate to $\mathcal{C}_{\mathrm{H}}$. Part (i) confines the residual error to the unacquired samples. The factor of two comes from the nonconvex rank constraint: Best approximation onto $\mathcal{C}_{\mathrm{H}}$ is not firmly nonexpansive, unlike projection onto a convex consistency set. In practice, Eq.~\eqref{eq:hankel_projection} approximates this projection and enforces exact data consistency at every final replacement step.

Algorithm~\ref{alg:kgr_lrgr} summarizes the full reconstruction procedure. In the algorithm, $\Pi_{\mathrm{H}}^{\mathrm{init}}$ and $\Pi_{\mathrm{H}}^{\mathrm{final}}$ denote the same Hankel compatibility projection implemented with the initialization and final-projection settings, respectively, and $N_{\mathrm{KGR}}$ is the number of KGR fitting iterations.

\begin{algorithm}[t]
\caption{KGR Reconstruction Procedure}
\label{alg:kgr_lrgr}
\begin{algorithmic}[1]
\REQUIRE $\hat{K}$, $M_{\Omega}$, $\{\mathcal{B}_r\}_{r=1}^4$, $\{\mathcal{J}_r\}_{r=1}^4$, $\lambda_{\Omega}$, $\delta_{\mathrm{max}}$, $N_{\mathrm{KGR}}$, settings for $\Pi_{\mathrm{H}}^{\mathrm{init}}$ and $\Pi_{\mathrm{H}}^{\mathrm{final}}$
\STATE \textbf{Stage 1: Structured low-rank initialization}
\STATE Compute $K_{\mathrm{init}} = \Pi_{\mathrm{H}}^{\mathrm{init}}(\hat{K})$ with acquired-sample replacement.
\STATE \textbf{Stage 2: Frequency-adaptive KGR fitting}
\FOR{$r=1$ to $4$}
    \IF{$r<4$}
        \STATE Initialize isotropic Gabor-Gaussian primitives in $\mathcal{B}_r$.
    \ELSE
        \STATE Initialize anisotropic Gabor-Gaussian primitives in the high-frequency band.
    \ENDIF
\ENDFOR
\FOR{$m=1$ to $N_{\mathrm{KGR}}$}
    \STATE Update KGR parameters $\theta$ by minimizing Eq.~\eqref{eq:kgr_loss}.
\ENDFOR
\STATE \textbf{Stage 3: Reliability-gated augmentation}
\FOR{$r=1$ to $4$}
    \STATE Compute $\delta_r$ using Eq.~\eqref{eq:band_residual}.
\ENDFOR
\STATE Form $K_{\mathrm{aug}}$ using Eq.~\eqref{eq:gated_fill}.
\STATE \textbf{Stage 4: Low-rank compatibility projection}
\STATE Compute $K_{\mathrm{final}} = \Pi_{\mathrm{H}}^{\mathrm{final}}(K_{\mathrm{aug}})$ with acquired-sample replacement.
\STATE \textbf{Return} $K_{\mathrm{final}}$
\end{algorithmic}
\end{algorithm}

\section{Experiments}

\subsection{Experimental Setup}

\textbf{Datasets:}
We evaluated KGR on two fully sampled multi-coil brain magnetic resonance MRI benchmarks with data from $10$ subjects, with $5$ subjects from each benchmark. FastMRI brain data \cite{knoll2020fastmri,knoll2020fastmriChallenge,muckley2021fastmriChallenge} contributes $100$ standardized slices ($50$ T1-weighted and $50$ T2-weighted), while Calgary-Campinas-359 (CC359) \cite{souza2018cc359}, a 12-coil brain dataset with a matrix $218\times170$, contributes $80$ slices. Both benchmarks provide complex multi-coil k-space signal, and every reconstruction is evaluated as a coil-combined root-sum-of-squares (RSS) magnitude image. Each slice is reconstructed under four undersampling masks, giving $400$ fastMRI cases and $320$ CC359 cases for aggregate evaluation.

\textbf{Compared methods:}
We compare KGR with five baselines spanning reconstruction assumptions: \textbf{CS-TV}, which performs compressed sensing with data consistency and total-variation regularization \cite{lustig2007sparse}; \textbf{SAKE}, a calibrationless block-Hankel structured low-rank completion \cite{shin2014sake}; \textbf{P-LORAKS}, a calibrationless structured low-rank reconstruction with parallel imaging data \cite{haldar2016ploraks}; \textbf{ZS-SSL}, a scan-specific self-supervised reconstruction that estimates coil sensitivity maps from the acquired autocalibration center and trains a deep unrolled network \cite{yaman2020ssdu}; and \textbf{Image-GR}, an in-house image-space Gaussian baseline. Image-GR fits Gaussian primitives in the spatial domain and applies the same structured low-rank refinement, serving as the image-domain counterpart to the proposed k-space representation. The result reported for KGR is the final output of Algorithm~\ref{alg:kgr_lrgr}, while the intermediate Gaussian-filled k-space is analyzed in the ablation study. All methods receive the same acquired measurements and are evaluated with the same RSS coil-combination and metric protocol.

\textbf{Implementation details:}
KGR follows Algorithm~\ref{alg:kgr_lrgr} using the symbols of Section~III. Initialization $\Pi_{\mathrm{H}}^{\mathrm{init}}$ uses a $6\times6$ block-Hankel window and a low-rank threshold of $1.5$ for $100$ iterations. The four radial bands receive $50/600/2000/10000$ primitives; the high-frequency band uses the anisotropic radial-tangential covariance of Eq.~\eqref{eq:gabor_atom} with widths $\sigma_{\mathrm{rad}}=3.0$ and $\sigma_{\mathrm{tan}}=12.0$. Parameters are optimized for $N_{\mathrm{KGR}}=6000$ iterations with Adam at learning rate $0.03$, acquired-sample anchor weight $\lambda_{\Omega}=0.1$, and reliability gate $\delta_{\mathrm{max}}=0.40$ in Eq.~\eqref{eq:gated_fill}. The final projection $\Pi_{\mathrm{H}}^{\mathrm{final}}$ of Eq.~\eqref{eq:lrgr_definition} uses a $10\times10$ window and threshold $2.0$ for $100$ iterations. On CC359, whose smaller matrix and twelve coils favor a sparser representation, we reduce the band counts to $10/80/300/1100$, use an $8\times8$ final-projection window with threshold $1.8$, and set $\delta_{\mathrm{max}}=0.70$; all other settings remain unchanged.

For CS-TV, sensitivity maps are estimated from a $24\times24$ autocalibration signal (ACS) region, and a fixed isotropic-TV weight of $10^{-3}$ is used for $200$ ISTA iterations. P-LORAKS uses an exact-data-consistent $S$-matrix formulation with radius $2$ and rank $70$; its inner conjugate-gradient solver uses at most five iterations. Standalone SAKE uses the same $6\times6$ window, threshold $1.5$, and $100$ iterations, and ZS-SSL is trained per scan for up to $300$ epochs with Adam. All experiments are conducted on an NVIDIA GeForce RTX 5090 GPU.

\textbf{Evaluation metrics:}
All metrics are computed on coil-combined RSS magnitude images. The reconstruction and reference are normalized by the maximum reference intensity. We report PSNR, SSIM, and NMSE under a unified evaluation protocol. Qualitative comparisons use matched display windows, zoomed regions of interest, and error maps. The source code of KGR and experimental results are available for reproduction at: \url{https://github.com/yqx7150/KGR}.

\subsection{Experimental Results}

\textbf{Quantitative comparison:}
Tables~\ref{tab:fastmri_quantitative} and~\ref{tab:cc359_quantitative} report the main quantitative comparison on fastMRI and CC359 benchmarks under the primary Poisson and radial undersampling settings, evaluated separately at two acceleration factors ($R=6$ and $R=10$). Each setting averages $100$ fastMRI slices and $80$ CC359 slices. For each baseline, the $p$-value column reports a two-sided paired $t$-test against KGR on PSNR averaged over the selected slices and the four sampling settings.

\begin{table*}[!t]
\centering
\caption{Quantitative Comparison on the fastMRI Benchmark Under Poisson and Radial Undersampling at Acceleration Factors $R=6$ and $R=10$, Reporting PSNR, SSIM, NMSE, and Paired $t$-Test Significance}
\label{tab:fastmri_quantitative}
\setlength{\tabcolsep}{4pt}
\renewcommand{\arraystretch}{1.15}
\resizebox{\textwidth}{!}{%
\begin{tabular}{l*{13}{c}}
\hline
\multirow{2}{*}{Method} & \multicolumn{3}{c}{Poisson R6} & \multicolumn{3}{c}{Radial R6} & \multicolumn{3}{c}{Poisson R10} & \multicolumn{3}{c}{Radial R10} & \multirow{2}{*}{$p$-value} \\
\cline{2-4}\cline{5-7}\cline{8-10}\cline{11-13}
 & PSNR$\uparrow$ & SSIM$\uparrow$ & NMSE$\downarrow$ & PSNR$\uparrow$ & SSIM$\uparrow$ & NMSE$\downarrow$ & PSNR$\uparrow$ & SSIM$\uparrow$ & NMSE$\downarrow$ & PSNR$\uparrow$ & SSIM$\uparrow$ & NMSE$\downarrow$ & \\
\hline
Zero-fill & 30.63 & 0.8506 & 0.0193 & 29.58 & 0.8261 & 0.0251 & 28.71 & 0.7928 & 0.0310 & 26.88 & 0.7454 & 0.0459 & $1.1\times10^{-8}$ \\
CS-TV~\cite{lustig2007sparse} & 33.49 & 0.9009 & 0.0094 & 33.15 & 0.8914 & 0.0101 & 33.49 & 0.8835 & 0.0097 & 32.59 & 0.8722 & 0.0114 & $7.4\times10^{-6}$ \\
SAKE~\cite{shin2014sake} & 34.89 & 0.9036 & 0.0066 & 34.32 & 0.8913 & 0.0074 & 33.13 & 0.8752 & 0.0099 & 32.72 & 0.8669 & 0.0107 & $6.2\times10^{-7}$ \\
P-LORAKS~\cite{haldar2016ploraks} & 34.16 & 0.9007 & 0.0077 & 33.43 & 0.8882 & 0.0091 & 32.39 & 0.8677 & 0.0116 & 31.39 & 0.8497 & 0.0148 & $8.7\times10^{-7}$ \\
Image-GR & 33.25 & 0.8589 & 0.0105 & 32.89 & 0.8585 & 0.0113 & 32.27 & 0.8381 & 0.0128 & 32.05 & 0.8450 & 0.0133 & $1.1\times10^{-6}$ \\
ZS-SSL~\cite{yaman2020ssdu} & 33.26 & 0.8918 & 0.0104 & 32.89 & 0.8802 & 0.0112 & 32.59 & 0.8661 & 0.0120 & 32.29 & 0.8612 & 0.0127 & $6.3\times10^{-6}$ \\
KGR & \textbf{35.69} & \textbf{0.9148} & \textbf{0.0054} & \textbf{35.18} & \textbf{0.9000} & \textbf{0.0059} & \textbf{33.80} & \textbf{0.8891} & \textbf{0.0082} & \textbf{33.41} & \textbf{0.8739} & \textbf{0.0091} & --- \\
\hline
\end{tabular}%
}
\end{table*}

\begin{table*}[!t]
\centering
\caption{Quantitative Comparison on the CC359 Benchmark Under Poisson and Radial Undersampling at Acceleration Factors $R=6$ and $R=10$, Reporting PSNR, SSIM, NMSE, and Paired $t$-Test Significance}
\label{tab:cc359_quantitative}
\setlength{\tabcolsep}{4pt}
\renewcommand{\arraystretch}{1.15}
\resizebox{\textwidth}{!}{%
\begin{tabular}{l*{13}{c}}
\hline
\multirow{2}{*}{Method} & \multicolumn{3}{c}{Poisson R6} & \multicolumn{3}{c}{Radial R6} & \multicolumn{3}{c}{Poisson R10} & \multicolumn{3}{c}{Radial R10} & \multirow{2}{*}{$p$-value} \\
\cline{2-4}\cline{5-7}\cline{8-10}\cline{11-13}
 & PSNR$\uparrow$ & SSIM$\uparrow$ & NMSE$\downarrow$ & PSNR$\uparrow$ & SSIM$\uparrow$ & NMSE$\downarrow$ & PSNR$\uparrow$ & SSIM$\uparrow$ & NMSE$\downarrow$ & PSNR$\uparrow$ & SSIM$\uparrow$ & NMSE$\downarrow$ & \\
\hline
Zero-fill & 25.83 & 0.7561 & 0.0307 & 25.08 & 0.7231 & 0.0365 & 24.34 & 0.6730 & 0.0433 & 22.90 & 0.5913 & 0.0596 & $4.0\times10^{-12}$ \\
CS-TV~\cite{lustig2007sparse} & 28.49 & 0.8255 & 0.0162 & 27.77 & 0.8035 & 0.0199 & 27.55 & 0.7961 & 0.0208 & 25.51 & 0.7217 & 0.0335 & $2.6\times10^{-10}$ \\
SAKE~\cite{shin2014sake} & 30.31 & 0.8674 & 0.0102 & 29.58 & 0.8579 & 0.0124 & 28.69 & 0.8314 & 0.0153 & 26.94 & 0.7837 & 0.0236 & $2.0\times10^{-7}$ \\
P-LORAKS~\cite{haldar2016ploraks} & 29.87 & 0.8681 & 0.0121 & 28.57 & 0.8408 & 0.0168 & 27.97 & 0.8181 & 0.0192 & 25.89 & 0.7458 & 0.0315 & $9.8\times10^{-7}$ \\
Image-GR & 28.88 & 0.8438 & 0.0142 & 28.30 & 0.8289 & 0.0167 & 27.59 & 0.8003 & 0.0192 & 25.79 & 0.7281 & 0.0303 & $6.9\times10^{-9}$ \\
ZS-SSL~\cite{yaman2020ssdu} & 29.89 & 0.8881 & 0.0119 & 28.84 & 0.8644 & 0.0155 & 28.28 & 0.8379 & 0.0176 & 26.74 & 0.7796 & 0.0264 & $4.4\times10^{-6}$ \\
KGR & \textbf{31.23} & \textbf{0.8931} & \textbf{0.0083} & \textbf{30.24} & \textbf{0.8752} & \textbf{0.0108} & \textbf{29.39} & \textbf{0.8528} & \textbf{0.0130} & \textbf{27.35} & \textbf{0.7984} & \textbf{0.0214} & --- \\
\hline
\end{tabular}%
}
\end{table*}

Across all eight dataset-by-mask settings, KGR attains the best PSNR, SSIM, and NMSE among all compared methods. Paired $t$-tests show that KGR significantly outperforms every baseline on PSNR in both datasets. Compared with standalone SAKE, KGR improves PSNR by up to about $0.9$ dB and also improves SSIM and NMSE. Image-GR applies the same low-rank refinement to an image-domain Gaussian fit. Its lower scores, about $2.4$ dB PSNR below KGR on fastMRI Poisson R6, isolate the role of placing the Gaussian representation in k-space. KGR also outperforms the self-supervised ZS-SSL in every setting. ZS-SSL estimates coil sensitivity maps from the autocalibration center and trains a deep unrolled network. KGR is calibrationless and network-free. The margin is largest under Poisson R6 and narrows toward radial R10.

\begin{figure*}[!t]
    \centering
    \includegraphics[width=\textwidth]{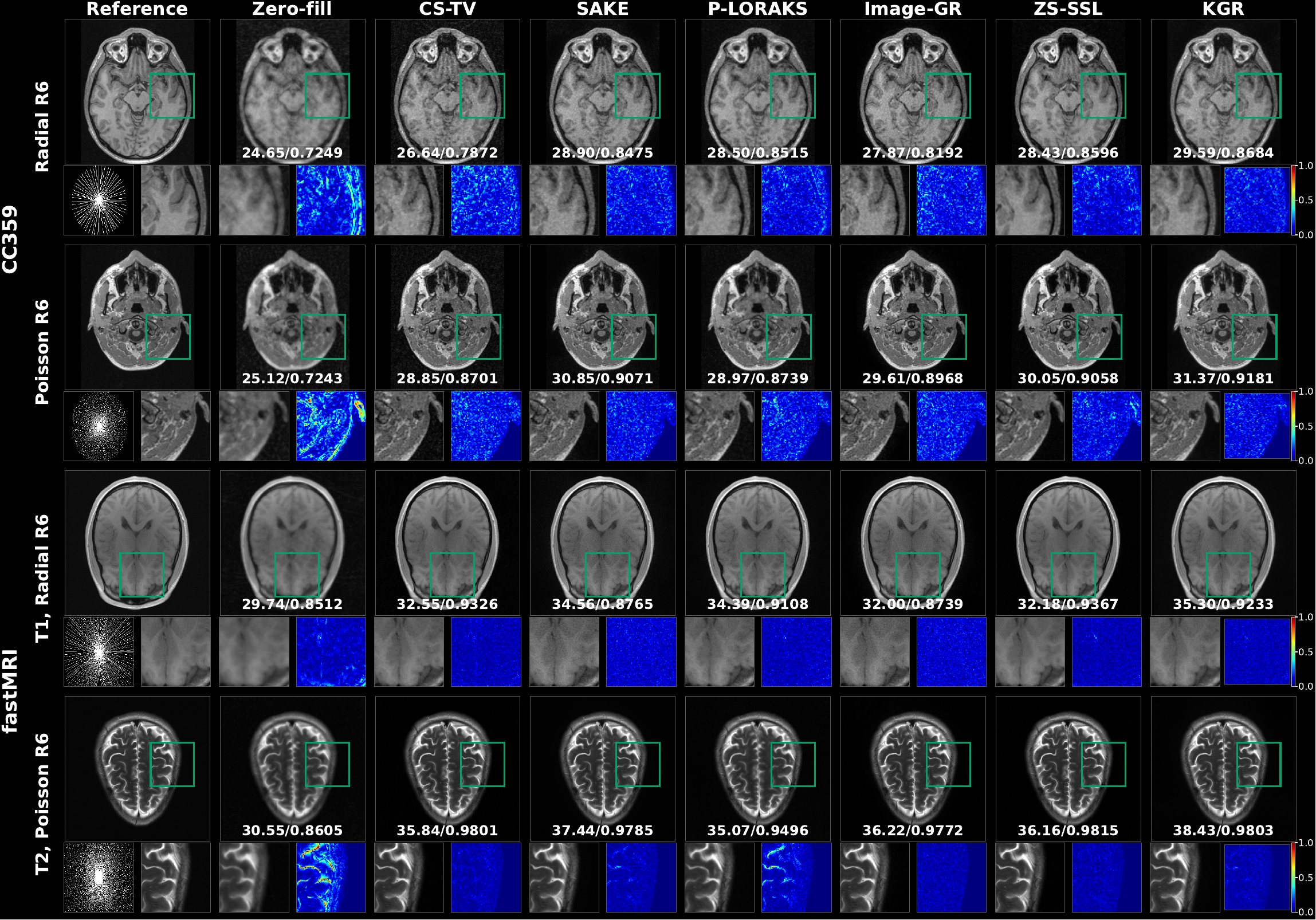}
    \caption{Qualitative comparison on fastMRI and CC359 under Poisson R6 and radial R6 undersampling. Columns show the fully sampled reference, zero-fill, CS-TV, SAKE, P-LORAKS, Image-GR, ZS-SSL, and KGR. The top two examples are from CC359, and the bottom two are from fastMRI. Each case pairs the full RSS reconstruction with the undersampling mask, the green-boxed ROI, and the $5\times$-amplified ROI error map within the reference-derived brain ROI. Displayed values are PSNR and SSIM, with a matched intensity window across methods.}
    \label{fig:qualitative_combined}
\end{figure*}

\textbf{Qualitative comparison:}
Fig.~\ref{fig:qualitative_combined} shows representative reconstructions on both datasets under the primary undersampling settings, following the column order and display protocol specified in the figure caption. Consistent with Tables~\ref{tab:fastmri_quantitative} and~\ref{tab:cc359_quantitative}, zero-fill retains strong undersampling artifacts. The structured low-rank baselines reduce most aliasing but leave residual local errors along anatomical boundaries. KGR reduces the ROI error patterns and preserves fine cortical structure in the highlighted regions. The comparison shows representative behavior under the displayed masks and should not be read as a population-level visual claim.

\subsection{Ablation Study}

To examine the contribution of each design component, we performed a leave-one-out ablation in $100$ CC359 cases. The cases comprise $50$ slices evaluated under both Poisson R6 and radial R6 undersampling. Starting from the full method, we evaluate variants that omit the frequency-adaptive decomposition in Eq.~\eqref{eq:kgr_loss} and fit a single band over the full grid, skip the final low-rank compatibility projection $\Pi_{\mathrm{H}}^{\mathrm{final}}$ of Eq.~\eqref{eq:lrgr_definition}, or replace directional Gabor modeling by dropping the Gabor carrier $\boldsymbol{\xi}_j$ and the anisotropic radial-tangential covariance of Eq.~\eqref{eq:gabor_atom}. In the last variant, each high-frequency primitive is reduced to an isotropic Gaussian envelope. For each variant, we also report a two-sided paired $t$-test on the paired PSNR differences per-case.

Table~\ref{tab:mechanism_ablation} shows that every component contributes to the final reconstruction. The full method achieves the best PSNR, SSIM, and NMSE. Frequency-adaptive band decomposition provides the largest benefit: Its omission causes a $2.48$ dB PSNR drop, a $0.060$ SSIM drop, and the worst NMSE. The final low-rank compatibility projection contributes the next largest improvement, with a $0.86$ dB PSNR gain relative to its ablated variant and corresponding gains in SSIM and NMSE. Directional Gabor modeling adds a smaller but significant refinement. These results support the use of both the continuous Gabor-Gaussian representation and the structured low-rank refinement in the tested CC359 cases and sampling patterns.

\begin{table}[!htbp]
\centering
\caption{Ablation Study of the Proposed Method on CC359}
\label{tab:mechanism_ablation}
\small
\setlength{\tabcolsep}{2.5pt}
\renewcommand{\arraystretch}{1.08}
\begin{tabular}{@{}lcccc@{}}
\hline
Configuration & PSNR $\uparrow$ & SSIM $\uparrow$ & NMSE $\downarrow$ & $p$-value \\
\hline
Full method & \textbf{30.92} & \textbf{0.8855} & \textbf{0.0098} & --- \\
w/o freq.-adaptive & 28.45 & 0.8259 & 0.0177 & $1.9\times10^{-6}$ \\
w/o LR projection & 30.06 & 0.8738 & 0.0122 & $5.7\times10^{-6}$ \\
w/o directional Gabor & 30.50 & 0.8784 & 0.0108 & $1.7\times10^{-5}$ \\
\hline
\end{tabular}

\end{table}

\begin{figure}[!htbp]
    \centering
    \includegraphics[width=\columnwidth]{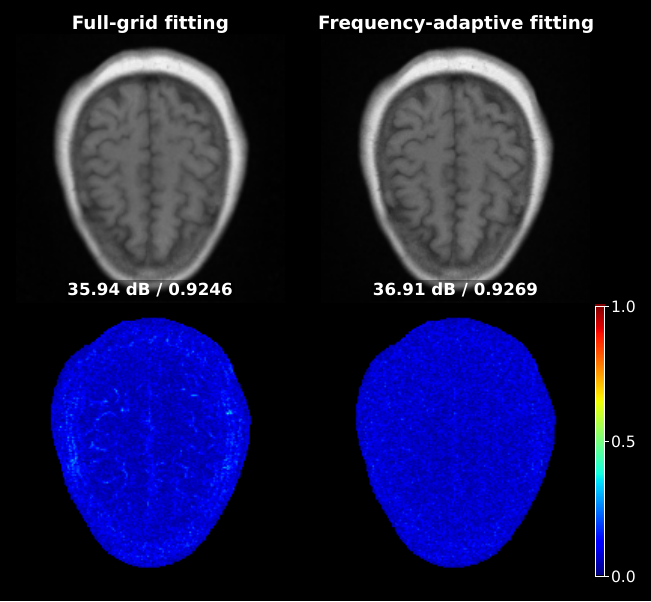}
    \caption{Qualitative effect of frequency-adaptive fitting on a representative fastMRI T1 slice under Poisson R6.}
    \label{fig:frequency_adaptive_ablation}
\end{figure}

Fig.~\ref{fig:frequency_adaptive_ablation} compares the full-grid and frequency-adaptive fitting on a representative fastMRI T1 slice under Poisson R6. Both variants use identical initialization, low-rank projection, and output processing, thus isolating the effect of the fitting strategy. Frequency-adaptive fitting reduces the distributed reconstruction residuals and improves PSNR/SSIM from $35.94$ dB/$0.9246$ to $36.91$ dB/$0.9269$.

\section{Discussion}
The results clarify how KGR complements structured low-rank reconstruction. KGR explicitly parameterizes local frequency-domain variation with Gabor-Gaussian primitives, whereas the Hankel projections enforce low-rank compatibility before and after continuous fitting. The ablation study supports this division of roles: Frequency-adaptive KGR fitting provides the largest gain, the final low-rank compatibility projection gives the next largest improvement, and directional Gabor modeling provides a smaller refinement. KGR therefore acts as a continuous representation module that augments, rather than replaces, structured low-rank reconstruction.

Consistent performance across Poisson and radial sampling suggests that KGR is not tied to a single gap geometry. Poisson sampling creates irregular missing locations, whereas radial sampling produces directional gaps. Improvements under both patterns indicate that the representation can accommodate distinct k-space missing-data structures. This evidence supports a focused claim: KGR complements structured low-rank reconstruction under the irregular and directional sampling patterns evaluated here.

The comparison with Image-GR provides evidence for placing the Gaussian representation in native k-space. Both methods use the same low-rank compatibility projection, but KGR performs better across all evaluated settings, supporting direct modeling of complex multi-coil k-space rather than applying Gaussian primitives after image formation. Shared geometry captures common frequency-domain support, while coil-specific complex amplitudes preserve receiver-dependent modulation. Like calibrationless structured low-rank methods \cite{shin2014sake}, KGR is scan-specific and requires neither population-level pretraining nor explicit coil sensitivity maps.

Several limitations define directions for future work. The current implementation is scan-specific and two-dimensional, and its robustness to uniform Cartesian undersampling, particularly at high acceleration, remains limited. Combining KGR with calibration-based k-space interpolation may improve this setting. Although the continuous representation can be evaluated off grid, extension to true non-Cartesian, volumetric, or dynamic acquisitions will require trajectory-aware data consistency and corresponding low-rank operators.

\section{Conclusion}

We introduce KGR, an explicit continuous representation framework that addresses a fundamental limitation of existing parallel MRI reconstruction methods: The absence of continuous modeling in native k-space signal. By parameterizing k-space as a continuous signal and integrating it with structured low-rank refinement, KGR bridges local frequency-domain modeling with intrinsic multi-coil redundancy, enabling reconstruction directly from measurements without relying on pretrained network. Extensive experiments demonstrate that KGR consistently improves reconstruction performance across different datasets and under-sampling patterns, validating the complementary advantages of continuous signal representation and physics-driven k-space constraints. This work highlights explicit continuous k-space modeling as a promising direction for developing more flexible and physically grounded reconstruction methods for accelerated parallel MRI.

\section*{REFERENCES}
\bibliographystyle{ieeetr}
\bibliography{ref} % force rebuild

\end{document}